\documentclass[runningheads]{llncs}

\usepackage{algorithm}
\usepackage[noend]{algpseudocode}
\usepackage{graphicx}

\usepackage{amsmath}
\DeclareMathOperator{\Viterbi}{Viterbi}
\DeclareMathOperator*{\MDPVI}{MDP}

\begin{document}
\title{Markov Decision Process for MOOC users behavioral inference\thanks{Supported by ANR-15-IDFN-0003-04.}}
%
%
\author{Firas JARBOUI\inst{1,2} \and C\'elya GRUSON-DANIEL\inst{3,5} \and Pierre Chanial\inst{1} \and
Alain DURMUS\inst{2} \and Vincent ROCCHISANI\inst{1} \and Sophie-Helene GOULET EBONGUE\inst{3} \and Anneliese DEPOUX\inst{3,4} \and Wilfried KIRSCHENMANN\inst{1} \and 
Vianney PERCHET\inst{2}}
\authorrunning{F. JARBOUI et al.}
\institute{ANEO, Boulogne Billancourt, France \and CMLA, \'Ecole normale sup\'erieur Paris Saclay, Universit\'e Paris Saclay, France \and Centre Virchow-Villerm\'e for Public Health Paris-Berlin, Universit\'e Sorbonne Paris-Cit\'e, France, \and GRIPIC - EA 1498, , Sorbonne Universit\'e, France, \and DRISS (Digital Research in Science \& Society) }
\maketitle              
%
\begin{abstract}
Studies on massive open online courses (MOOCs) users discuss the existence of typical profiles and their impact on the learning process of the students. However defining the typical behaviors as well as classifying the users accordingly is a difficult task. In this paper we suggest two methods to model MOOC users behaviour given their log data. We mold their behavior into a Markov Decision Process framework. We associate the user's intentions with the MDP reward and argue that this allows us to classify them.

\keywords{User Behaviour Studies  \and Learning Analytics \and Markov Decision Process \and Inverse Reinforcement Learning.}
\end{abstract}
\section{Introduction}

Finding an efficient way to identify behavioural patterns of MOOC users community is a recurring issue in e-learning. However, as detailed in the review of Romero and Ventura \cite{ref_article1} on educational data science research, the way this problem was studied was either testing correlations given conjectures or trying to identify communities of look-a-likes.

The main approaches study aggregates of data generated by users in order to identify their respective behaviors with respect to some typology of the students. 
For instance, Ramesh A. et al. \cite{ref_article2} distinguish learner behavior according to their engagement into \textit{active}, \textit{passive} and \textit{disengaged} learners. They predict their behavior based on a probabilistic soft logic model taking into account users features relating to their engagement on the Mooc. Corin L. et al. \cite{ref_article3} describes the behaviour of users through flow diagrams of the state transitions through comparing different sets of behaviours with graphical models. Cheng Y. and Gautam B. \cite{ref_article4} predicted users drop-out through temporal granularities in features suspected to influence the drop-out of the users. Finally, Christopher B. et al. \cite{ref_article5} cover the prediction of the users achievement through their levels of activity in the course.
Unfortunately, given a different typology of users profile, we can not easily transpose these approaches because the features used to characterize the learner are selected with respect to the definition of the classes. For example, performance related typology can be matched to users' quizzes success rate, the drop-out rate oriented classes can be tracked through the times series of connection history. However the task becomes more difficult when the definition of classes can not be reduced to a simple quantifiable measurement, for example the user intentions, goals and motivations.
Chase G. and Cheng Z. \cite{ref_article6} aimed for a generalized method by modelling the user behaviour as a two layer hidden Markov model. They used log data to construct their characterization of the users. They cluster the transition probabilities to define the hidden states, and they compare the transitions between these hidden states for the high and low performing students. However, if we only observe the transitions probabilities, identifying an interpretation of the associated behavior is a complex task.

We aim to define general models to study any kind of user behavior without loosing the interpretability of results. We consider two different main models. In the first one, we assume that each user can adopt his own different policy. Each user has a reward function over the MOOC and tries to optimize it. We can cluster theses rewards into a finite number of classes that represent the behaviors explaining the observations. 
In the second one, we assume that there are a limited number of rewards the users can optimize. Each reward translates into a typical behavior and the users are switching between them along the MOOC.

\section{Mathematical Preliminary}

\subsection{Markov Decision Process}
Consider a Markov Decision Process (MDP) $\mathcal{M}=\{\mathcal{S}, \mathcal{A}, \mathcal{P}, \mathcal{R}, \nu, \mathcal{X} \}$ where $\mathcal{S} = \{ 1,2, ..., N_s\} $ is a finite state space, $\mathcal{A} = \{1,2,...,N_a \}$ is a finite action space, $\mathcal{P}: \mathcal{S}\times\mathcal{A}\times\mathcal{S} \rightarrow [0,1]$ is the probability distribution of state transition such as $\mathcal{P}(s,a,s') = \mathrm{P}(s_{t+1}=s'|s_t=s,a_t=a)$ is a probability of going to state $s'$ under action $a$ from state $s$ , $\mathcal{R}: \mathcal{S}\times\mathcal{A} \rightarrow [0,1]$ is the reward function, $\nu \in ]0,1[$ is a discount factor and $\mathcal{X}$ is an initial distribution over the states i.e. $\mathrm{P}(s_0=s) = \mathcal{X}(s)$.
Any transition matrix compatible with the MDP on $\mathcal{S}\times\mathcal{A}$ is referred to as a policy, and we denote $\pi: \mathcal{S}\times\mathcal{A} \rightarrow [0,1]$ such policy. An agent following the policy $\pi$ would take at time $t$ the action $a_t=a$ at state $s_t=s$ with probability $P(a_t=a|s_t=s) = \pi(s,a)$. The value of the policy $\pi$ is $V^{\pi} = \sum_s \mathcal{X}(s) V^{\pi}(s)$ where $V^{\pi}(s) = \mathbf{E}^{\pi}[ \sum_t \nu^t \mathcal{R}(s_t,a_t) | s_0 = s]$ is the state value function.
Similarly we denote the state-action value function by $Q^{\pi}(s,a)$ defined by $ Q^{\pi}(s,a) = \mathbf{E}^{\pi}[ \sum_t \nu^t \mathcal{R}(s_t,a_t) | s_0 = s, a_0=a] $.
We denote by $\pi^*$ the optimal policy maximizing the expected discounted reward given any starting state, and by $V^*$ and $Q^*$ the corresponding values defined as \cite{ref_article7}: 
\[ 
\begin{array}{rcll}
    \pi^* = \mathrm{argmax}_{\pi} V^{\pi}(s) & = & \mathrm{argmax}_{\pi} Q^{\pi}(s,a) & \; \forall s\in\mathcal{S} \; \forall a\in\mathcal{A} \\
    V^*(s) & = & V^{\pi^*}(s) & \; \forall s\in\mathcal{S}  \\
    Q^*(s,a) & = & Q^{\pi^*}(s,a) & \; \forall s\in\mathcal{S} \; \forall a\in\mathcal{A}.
\end{array}
 \]

Given an MDP $\mathcal{M}$, the optimal policy $\pi^*$ can be computed using well-known methods such as Value iteration or Policy iteration \cite{ref_article7}. We use a Value iteration approach and we refer to this procedure as $Q^*=\MDPVI(\mathcal{M})$ 

\textbf{Reward parametrized MDP:}
In this paper we consider the case of MDPs with linearly parametrized rewards $\mathcal{M}_{\theta} = \{\mathcal{S}, \mathcal{A}, \mathcal{P}, \mathcal{R}_\theta, \phi, \nu, \mathcal{X} \} $ i.e. such that: 
\[\forall s\in\mathcal{S} \; \forall a\in\mathcal{A} \; \mathcal{R}_\theta(s,a) = \theta.\phi(s,a)^T,\]
where $\phi: \mathcal{S}\times\mathcal{A} \rightarrow \mathbf{R}^N$ is a feature map from the state-action space to a real valued N dimensional space and $\theta \in \mathbf{R}^N$ is an N dimensional real weights.

For a given $\theta$, we will denote the optimal policy by $\pi^*_\theta$, and the corresponding optimal value functions by $V^*_\theta$ and $Q^*_\theta$ .

\subsection{Inverse Reinforcement Learning}
Let be $\mathcal{M}_{\theta}$ a reward paramatrized MDP whose parameter $\theta$ is unknown. We denote by $\mathcal{W} = \{\mathcal{S}, \mathcal{A}, \mathcal{P}, \phi, \nu, \mathcal{X} \}$, and $\mathcal{D}_M = \{ (y^i_{0:T_i})_{i=1}^M\}$ the behavioral data where $y^i_{0:T_i} = \{ y_t^i=(s_t^i,a_t^i)_{i=1}^{T_i} \}$ are the $i^{th}$ individual following an unknown policy $\pi^*_{\theta}$. The goal of the Inverse Reinforcement Learning (IRL) problem is to identify parameters $\hat{\theta}$ such that $\pi^*_{\hat{\theta}}$ are as likely as $\pi^*_{\theta}$ to generate the observations  $\mathcal{D}_M$.  
The IRL problem is ill-posed \cite{ref_article8} as there exists infinitely many reward parameters that yield $\pi^*_{\theta}$ as an optimal policy. For example with $\theta = 0$, any policy is optimal for any IRL problem. 

To circumvent this issue, many approaches have been proposed to define preferences over the reward space. These approaches can be broadly divided in two settings: Optimization IRL and Bayesian IRL. Optimization oriented approaches define objective function that encode such preferences \cite{ref_article8,ref_article10,ref_article11}. Bayesian approaches formulate the reward preferences in the form of a prior distribution over the rewards and define behavior compatibility as a likelihood function \cite{ref_article12,ref_article14,ref_article15}.
We will follow the latter setting \cite{ref_article15}.
The model we consider assumes that that agents are not following an optimal policy $\pi^*_\theta$ but rather an aproximal one. more precisely, we assume that: 
\[ 
\left.
\begin{array}{lcll}
    \tilde{\pi}^\eta_\theta (s,a) & = & \frac{\mathrm{exp}( \eta Q^*_\theta(s,a) )}{ \sum_{a_i} \mathrm{exp}( \eta Q^*_\theta(s,a_i) }  &\; \forall s\in\mathcal{S} \; \forall a\in\mathcal{A}  \\
    P(a_t=a|s_t=s, \mathcal{M}_\theta) & = & \tilde{\pi}^\eta_\theta (s,a) & \; \forall s\in\mathcal{S} \; \forall a\in\mathcal{A} \\
    P(s_{t+1}=s' | s_t=a,a_t=a,  \mathcal{M}_{\theta}) & = & \mathcal{P}(s,a,s') &\; \forall s,s'\in\mathcal{S} \; \forall a\in\mathcal{A} \\
\end{array} 
\right. 
\]
Therefore, under this model, the likelihood is given by:
\[
\left.
\begin{array}{llcl}
& P(\mathcal{D}_M | \mathcal{M}_\theta) & = & \prod_{i=1}^M \prod_{t=1}^{T_i} \tilde{\pi}^\eta_\theta (s,a) \times \prod_{i=1}^M \prod_{t=1}^{T_i} P(s_{t+1} | s_t,a_t,  \mathcal{M}_\theta)  \\
\end{array}
\right.
\]
We have assumed that $(a_t,s_t)_{t\geq 0}$ is still Markovian, with transition probabilities given by $\mathcal{P}$ which does not depends on $\theta$. Thus, $\prod_{i=1}^M \prod_{t=1}^{T_i} P(s_{t+1} | s_t,a_t,  \mathcal{M}_\theta)$ can be treated as a multiplicative constant with respect to $\theta$. We define:
\begin{equation*}
\mathcal{L}(\theta; (\mathcal{D}_M, \mathcal{W})) = \prod_{i=1}^M \prod_{t=1}^{T_i} \frac{\mathrm{exp}( \eta Q^*_\theta(s_t^i,a_t^i) )}{ \sum_a \mathrm{exp}( \eta Q^*_\theta(s_t^i,a) },
\end{equation*}

where $\eta$ can be interpreted as a confidence parameter. The bigger it gets, the closer are the policies $\tilde{\pi}^\eta_\theta$ and $\pi^*_\theta$, as $ \mathrm{lim}_{\eta\rightarrow\infty} \tilde{\pi}^\eta_\theta (s,a)  = \pi^*_\theta $.
The posterior distribution is given by Bayes Theorem, where we choose $\theta\rightarrow\mathrm{P}(\theta)$ to be a uniform distribution over a subset of the parameter space.
\[ \mathrm{P}(\theta|\mathcal{D}_M) \;\alpha\; \mathcal{L}(\theta; (\mathcal{D}_M, \mathcal{W})) \mathrm{P}(\theta)\]

We use approximate samples from the distribution $\mathrm{P}(\theta|\mathcal{D}_M)$ to compute the a posteriori mean or median which are optimal under the square or linear loss function respectively \cite{ref_article15,ref_article18}. Iterating Algorithm \ref{theta} generates the samples.

\vspace{-.3cm}
\begin{algorithm}
\caption{SampleTheta}\label{theta}
\begin{algorithmic}[1]
\Procedure{$ \theta = \mathrm{SampleTheta}(\theta^0, \mathcal{W}, P, \sigma, \eta, \mathcal{D}_M)  $}{}
\State $\mathrm{sample } \epsilon \sim \mathcal{N}(0,1)$ and set $\tilde{\theta} = \theta^0 + \sigma\epsilon $ 
\State $\tilde{Q}^*_{\tilde{\theta}} = \MDPVI(\mathcal{M}_{\tilde{\theta}}) $
\State Set $ \theta = \tilde{\theta} $ with probability: 
      \State \hspace{2cm} $ \mathrm{min}(1, \frac{ \mathcal{L}(\mathcal{D}_M, \mathcal{M}_{\tilde{\theta}} P(\tilde{\theta}) } {\mathcal{L}(\mathcal{D}_M, \mathcal{M}_{\theta} P(\theta)} ) $
\State else set $\theta = \theta^0$
\State Return $\theta$
\EndProcedure
\end{algorithmic}
\end{algorithm}
\vspace{-1cm}
\subsection{Switched Markov Decision Process sMDP}
Inspired from switched Linear Dynamical Systems \cite{ref_article17}, switched Markov Decision Process allow us to simplify complex phenomena into transitions among a set of simpler models. For example, non-linear behaviors such as an individual's movement in a crowd, can be viewed as an array of linear behavior among which the person is temporally switching \cite{ref_article12}.

Let $(\mathcal{M}_{\theta_i})_{i=1}^L$ be a set of MDP models with corresponding parameters $(\theta_i)_{i=1}^L$ and policies $(\tilde{\pi}^\eta_{\theta_i})_{i=1}^L$. Switching between these models is governed by a discrete Markov process with transitions $\zeta$. 
We denote $ z_t $ as the latent mode of the system at time $t$, thus, it is sampled according to $\zeta_{z_{t-1},.}$. We also denote by $ y_t=(s_t,a_t) $ the observations which obey to a Markov decision process model.

\[
\left.
\begin{array}{rcl}
      z_t | \{ \zeta_i \}_{i=1}^L, z_{t-1} & \sim  & \zeta_{z_{t-1},.} \\
      y_t | y_{t-1},z_t, \{\theta_i\}_{i=1}^L  & \sim & \pi^\eta_{\theta_{z_t}}(y_t) \mathcal{P}(s_{t-1},a_{t-1},s_t)
\end{array}
\right.
\]

The hidden modes and the MDP associated with each one of them provide a Hidden Markov Model (HMM) structure leading to repeating simple behaviors. A common approach to solve HMM model is the forward backward method developed by Andrew Viterbi\cite{ref_article21}. We call $\Viterbi$ the procedure that evaluates the latent modes $\hat{z}_{1:T}$ and the transition probabilities $\mathcal{F}_{ij} = \mathrm{P}(z_t=i,z_{t+1}=j)$ and we denote it by $ [z_{1:T}, \mathcal{F}] = \Viterbi(y_{1:T}, \zeta_{kk'}, \pi_{\theta_i}^\eta) $.
We denote by $\mathcal{M}^S = (\mathcal{W}, L, (\theta_i)_{i=1}^L, \zeta, \mathcal{X}_m ) $ the sMDP model where $\mathcal{X}_m$ is the initial mode distribution.
As an application we will tackle the case where the parameters $ ((\theta_i)_{i=1}^L, \zeta) $ are unknown and will be learned from the data.

\subsection{Label Propagation}
Let $(x_i,y_i)_{i=1}^{l}$ be a set of labeled data, and $(x_i,y_i)_{i=l+1}^{l+u}$ be a set of unlabeled data, i.e. $(y_i)_{i=l+1}^{l+u}$ are unobserved. Where $ x_i \in \mathbf{R}^D $ and $ y_i \in \mathcal{C}$ for $i = 1,..,l+u$ and where $\mathcal{C} $ is a finite set. We define $X = \{ x_1, x_2, ... x_{l+u} \}$, $Y_L = \{ y_1, ..., y_l \}$ and $Y_U = \{ y_{l+1}, ..., y_{l+u} \}$. The problem is to estimate $Y_U$ given $X$ and $Y_L$. we also denote by $Y$ the matrix of label probability where $Y_{ic} = P(y_i = c)$.
We want to find a matrix Y that satisfies the following: 
\[
\left\{
\begin{array}{lcll}
    Y_{i.} & = & \frac{(TY)_{i.}}{||(TY)_{i.}||_2} \; & \forall i>l+1 \\
    Y_{i.} & = & \delta(y_i) & \forall i \leq l
\end{array}
\right.
\]
Where $T$ is the matrix of label transition probabilities through the set $X$. $T_{ij}$, the probability that $x_i$ will inherit the label of $x_j$, is proportional to the distances between the two points.
 Algorithm \ref{labelprop} \cite{ref_article20} solves for $Y$.

\[ 
\left.
\begin{array}{llclcl}
    & T_{ij} & = & P(y_j \rightarrow y_i) & = & \frac{w_{ij}}{\sum_{k=1}^{l+u}w_{kj}}   \\
    \mathrm{and} & w_{ij} & = & \mathrm{exp}(-\frac{d^2_{ij}}{\sigma^2}) & = & \mathrm{exp}(- \frac{|x_i - x_j|_2^2}{\sigma^2})
\end{array}
\right.
\]
\vspace{-.7cm}
\begin{algorithm}
\caption{Label Propagation}\label{labelprop}
\begin{algorithmic}[1]
\Procedure{$ Y = \mathrm{LabelProp}( X, Y_L, \sigma )  $}{}
\State initialize $Y$ with: $Y_{ic} = \delta(c = y_i) $ if $i \leq l$ and $Y_{ic} = \frac{1}{C}$ if $ i > l$
\State $repeat$ until convergence of $Y$:
\State \hspace{2cm} $Y = TY$
\State \hspace{2cm} row-normalize Y
\State \hspace{2cm} $Y_{ic} = \delta(c = y_i) $ for $i \leq l$
\State return $Y$
\EndProcedure
\end{algorithmic}
\end{algorithm}

\vspace{-1cm}
\section{Behavior inference with IRL}
We now suggest two ways to develop a classification for MOOC user behaviors. In the first one, \textbf{Static Behavior Clustering (SBC)},  we consider that each user follows a policy that optimizes his own reward function and that generates their log data. We use the reward parameter associated to each user as features and we propagate labels that experts define on a restricted set of users.
In the second one, \textbf{Dynamic Behavior Clustering (DBC)}, we consider that there is a small number of behaviors a user can adopt. We model their behaviors with a sMDP, the log data is then simplified into a set of typical behavior successions. 
We denote in the sequel by $\mathcal{W}$ the MDP associated to the MOOC and by $\mathcal{D}_M = \{ (y^i_{0:T_i})_{i=1}^M\} $ the collected observation from $M$ users where $y^i_t = (s_t^i,a_t^i)$ indicate user $i$ being at state $s_t^i$ at time $t$ and taking action $a_t^i$. We denote by $\mathcal{D}_M(i) = \{  y^i_{0:T_i} \}$ the data associated to the $i^{th}$ user.

\subsection{Construction of the MOOC MDP}
Given a MOOC, we first define some associated MDP parameters $\mathcal{W}$. The construction is straightforward: we define $\mathcal{S}$ as the different pages a user can access along with a resting state (associated to logging out of the website), $\mathcal{A}$ is associated to the different actions available to the user such as playing a video, clicking on a given link or answering a quiz. $\mathcal{X}$ and $\mathcal{P}$ are computed empirically given the data $\mathcal{D}_M$. 
The feature function $\phi$ however gives some flexibility to our approach. If we do not have much knowledge about the behaviors we are trying to track, we can define $\phi(s,a) = 1_{\mathcal{S}\times\mathcal{A}}(s,a)$ as the indicator function of each state-action combination. Unfortunately, this will become unhandy for higher dimensions as $\theta \in \mathbf{R}^{|\mathcal{S}|\times|\mathcal{A}|}$. An expert can however define a set of features to which the set of state-actions can be mapped. 
The discount parameter $\nu$ reflects the ability of the agents of long term planning. It should be learned along with other parameters as it might not be the same for each user.
However, we will consider a shared parameter that we fix at $0.9$ for the sake of simplicity.
 
\subsection{Static Behavior Clustering}
We assume the existence of a set of behavior classes $\mathcal{C} = \{1,..,N\}$. Let $(c_i)_{i=1}^M$ be the classes associated to each user in the data. With the help of a human expert, using highly restrictive conditions, we identify $l$ users classes. Without loss of generality, we assume that $C_L = \{c_1, ..., c_l\}$ is the set of known classes. Let $\mathbf{P}^c$ be the class probability matrix where $\mathbf{P}^c_{ij} = P(c_i = j)$
We consider also that each user behaves according to the MDP $\mathcal{M}_{\theta_i} = \{\mathcal{W}, \mathcal{R}_{\theta_i} \}$ 
such that: 
\[\forall t\leq T_i \; a_t^i \sim \pi^\eta_{\theta_i}(s_t^i,.) \; \mathrm{ and } \; s_{t+1}^i \sim \mathcal{P}(s_t^i,a_t^i,.) \]
For user $i$, we propose to infer such parameters given $\mathcal{D}_M(i)$, which will allow us to infer $(\theta_i)_{i=1}^M$. The objective is then to identify $\mathbf{P}^c$ given $\Theta$ and $C_L$. In Algorithm \ref{SBC} we suggest a method to solve this problem.
\vspace{-.5cm}
\begin{algorithm}
\caption{Static Behavior Clustering}\label{SBC}
\begin{algorithmic}[1]
\Procedure{$ \mathbf{P}^c = \mathrm{SBC}( \mathcal{D}_M, \mathcal{W}, C_L, P_\theta, \eta, \sigma_{MDP}, \sigma_{LP} )  $}{}
\State initialize $\Theta = \{ \theta_i \}$ to void values
\State $for$ i=1,..M:
\State \hspace{2cm} $\theta_i = BIRL(\mathcal{W}, P_\theta, \sigma, \eta, N_{\mathrm{max}}, \mathcal{D}_M(i)) $
\State $\mathbf{P}^c = \mathrm{LabelProp}(\Theta, C_L, \sigma_{LP})$
\State return $\mathbf{P}^c$
\EndProcedure
\end{algorithmic}
\end{algorithm}
\vspace{-1cm}
\subsection{Dynamic Behavior Clustering}
We suggest here a simpler version of the model of the sticky Hierarchical Dirichlet Process for sMDP \cite{ref_article12}. To simplify the problem we assume that the number of clusters $L$ is given. We suppose that the transition distribution from the $i^{th}$ mode are generated according to a Dirichlet distribution $\mathrm{Dir}(\alpha_L + \delta_i)$ where $\alpha_L = \alpha.1_L$, $\delta_i$ is the $i^{th}$ vector of the canonical base of $\mathbf{R}^L$, and $\alpha \in \mathbf{R}$. This avoids jumping between modes as the probability of remaining in the same mode is higher than switching to a new one. The reward parameter associated to each mode is sampled from $\mathrm{U}$ the uniform distribution over some subset of the parameter space. The observations obey to an sMDP model defined with $\mathcal{M}^S = (\mathcal{W}, L, \{ \theta_i, \zeta_i \}_{i=1}^L, \mathcal{X}_m ) $.
The full generative model is given bellow:
\[ 
\left.
\begin{array}{rcll}
    \zeta(i,.)|\alpha & \sim & Dir(\alpha_L+\delta_i) & \forall i=1,..,L \\
    \theta_i & \sim & \mathrm{U} & \forall i=1,..,L \\
    z_t^i|\{\zeta(i,.)\}_{j=1}^L & \sim & \zeta(z_{t-1},.) & \forall i=1,..,M \mathrm{ and } t=1,..T_i \\
    y_t^i|y_{t-1}^i, z_t^i, \{\theta_j\}_{j=1}^L & \sim & \pi^\eta_{\theta_{z_t^i}}(s_t^i,a_t^i)\mathcal{P}(s_{t-1}^i,a_{t-1}^i,s_t^i)  & \forall i=1,..,M \mathrm{ and } t=1,..T_i \\
\end{array}
\right.
\]

The intuition behind this model is that each user can adopt at each time step one of the behavior modes (i.e. he behaves accordingly to one of the MDPs).
To alleviate notations we denote in the following $ \Pi = \{\pi_{\theta_i}^\eta \}_{i=1}^L $, and $\mathcal{D}^k=\{y_t^i\; \forall i\leq L \; \forall t \leq T_i | z_t^i = k\}$ the observations associated to the $k^{th}$ behavior mode. 
We suggest a MCMC approach to solve this inference problem where each step looks as developed in Algorithm \ref{DBC}. We start by inferring the latent modes according to the previous parameters values. We define the new sample of the HMM parameters using the normalization of the frequencies probabilities $\mathcal{F}$ in step 6. In step 7, we split the data set according to the $z_t^i$ into $\mathcal{D}^i$ and use Algorithm \ref{theta} to sample the new MDP parameters $\theta_i$ and their policies.

\begin{algorithm}
\caption{Dynamic Behavior Clustering}\label{DBC}
\begin{algorithmic}[1]
\Procedure{$ \Theta^n, \zeta^n, \Pi^n  = \mathrm{DBC}( \mathcal{D}_M, \mathcal{W}, P_\theta, \eta, \sigma, \Theta^{n-1}, \zeta^{n-1}, \Pi^{n-1}, \alpha )  $}{}
\State Set $\mathcal{F} = [f_{ij}]_{i\in[1,L] \; j\in[1,L] } $ to zeros
\State $for$ i=1,..,M:
\State \hspace{2cm} $ [z^i_{1:T_i}, \mathcal{F}^i] = \Viterbi(y^i_{1:T_i}, \zeta^{n-1}, \Pi^{n-1})$
\State \hspace{2cm} $ \mathcal{F} = \mathcal{F} + \mathcal{F}^i $
\State $[\zeta^n] = \mathrm{SampleHMMParam}(\mathcal{D}_M, \mathcal{F}, \zeta^{n-1}, \alpha, \sigma)  $
\State $[\Theta^n, \Pi^n ] = \mathrm{SampleMDPParam}(\mathcal{D}_M, \Theta^{n-1}, z^i_{1:T_i}, \eta, \sigma) $
\State return $[\Theta^n, \zeta^n, \Pi^n]$
\EndProcedure
\end{algorithmic}
\end{algorithm}
\vspace{-1cm}
\section{Experiment}
The experiments were conducted on MOOCs published in the framework of the research project \#MOOCLive under the leadership of the Centre Virchow-Villerm\'e for Public Health. The project aimed to substantially improve the efficiency of the MOOCs through a deeper understanding of the participants and their behaviors. 
In the modelization of the MOOC's MDP, we used an indicator feature function over the state-action space $\mathcal{S}\times\mathcal{A}$.

\subsection{Static Behavior Clustering}
As mentioned before, our objective is to introduce a procedure that can be applied irrespectively of the experts classification. We experimented with multiple behavior classes $\mathcal{C}$. Each time, we first defined $\mathcal{C}$ and then identified the subset $C_L$.
For instance defining the \textit{collaborative behavior} as users who finish all quizzes and courses, get highest scores, and participate on the forum, the \textit{targeting behavior} which corresponds to a super student on only one chapter of the MOOC. Such perfect representation are rare, and only few users satisfy the required criterion. Afterward, we randomly select a testing set and ask experts either they agree or disagree with the classification. 
The experiment can have one of two possible outcomes, either we add the test set to the labeled set $C_L$ and run the algorithm again for better accuracy, or, we find that the expert becomes aware of some limitations and improves his behavior classes $\mathcal{C}$ definitions \cite{ref_article22}. This iterative process can be repeated as much as needed until the expert is satisfied with the outcome. In our case we converged to the following classification:
\begin{itemize}
    \item \textbf{Participant (P)} Does all the chapters, and answer all the quizzes;
    \item \textbf{Collaborative (C)} Does at least 70\% of all the chapters and quizzes but highly active on the forum;
    \item \textbf{Targeting (T)} Targets a chapter, solves the relevant quizzes;
    \item \textbf{Auditor (A)} Reads 70\% the chapters but answers $30\%$ of the quizzes;
    \item \textbf{Clicker (Cl)} Does not stay on the same page longer than 5 seconds;
    \item \textbf{Big Starter (BS)} Has a participant behavior up to the first 3 chapters; 
    \item \textbf{Late Quitter (LQ)} Has a participant behavior up to the last 3 chapters.
\end{itemize}

\subsection{Dynamic Behavior Clustering}
The results of \textbf{SBC} motivated the development of \textbf{DBC}. We observed that the classes that were satisfactory for the analysis requirement, are actually temporally characterized by a smaller number of simple behavior. \begin{itemize}
    \item \textbf{Exploration} where the user is randomly skipping through the MOOC;
    \item \textbf{Learning} where the user pays attention to the content of pages;
    \item \textbf{Certification} where the user is interested in the certification and tries to fulfill the courses requirements.
\end{itemize}

We considered a three dimensional feature space where the weight in each dimension reflects the probability of following the associated behavior. 
As expected, the users behaviors could be explained by the succession of simple behavior, we converged to 3 modes each of them optimizes one of the behaviors $\theta_1,\theta_2,\theta_3 \in \{[1,0,0],[0,1,0],[0,0,1]\}$.

\begin{figure}[h]
\hspace{.6cm}
\includegraphics[width=1\textwidth]{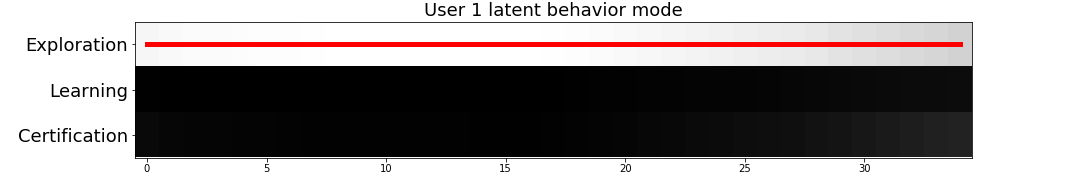}
\label{fig:VFF1}
\qquad
\includegraphics[width=1\textwidth]{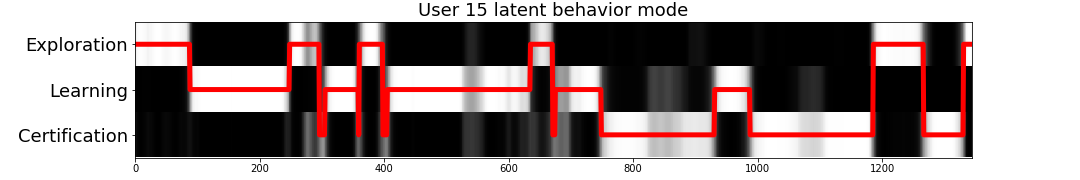}
\label{fig:VFF15}
\qquad
\includegraphics[width=1\textwidth]{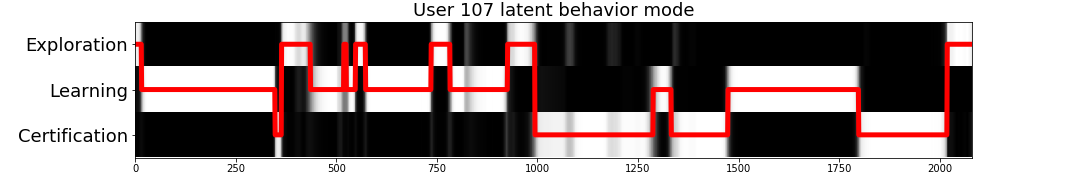}
\label{fig:VFF107}
\caption{Three \#MOOCLive users' latent behavior modes}
\label{fig:VFF}
\end{figure}

In Figure \ref{fig:VFF} we observe the expected temporal evolution of modes ($z^i_{t=0:T_i}$) for three users. 
Some behaviors such as the \textit{clicker} behavior, can be observed as an agent with the unique goal of exploration as shown in the case of user \#1. 
Other behaviors correspond to other patterns such as the \textit{late quitter}, whose behavior is quite similar to the \textit{participant}/\textit{collaborative} behavior. The user start exploring a little bit, oscillates between \textit{exploring} and \textit{learning} before adopting mostly the \textit{certifying} behavior. We end up sometimes with an exploration phase before leaving the courses entirely. 
The difference between the LQ and (C/P) behaviors is the length of the sequences. For instance we observe the fact that LQ tend to explore more by the end. 

\vspace{-1cm}

\section{Discussion}
From a practical point of view, both \textbf{SBC} and \textbf{DBC} were satisfactory as the results satisfied the experts who lead the experimentation process with us. We were able to improve our understanding of the users learning behaviors without requiring additional informations when treating different sets $\mathcal{C}$ of user classes. 
The results of the \textbf{SBC} are easily interpretable as the outputted behaviors are defined with the help of an expert. However, in the case of \textbf{DBC}, the task is more difficult. It mainly depends of the considered feature map and our ability to identify behaviors when observing the outputted parameters. In our cases, we did not struggle as we were anticipating such results. 

A big drawback of the \textbf{SBC} is the assumption that users behave according to a unique policy throughout the course. To circle around this, the behavior classes had to be specified enough to capture the nuances between the users. Even though \textbf{DBC} resolves partially this issue by allowing the users to jump among typical behaviors, a temporal explanation of the mode switching is far from being comprehensively satisfactory. In fact, the users are more likely to switch from an exploration behavior to a learning one because they visited a set of different pages (or states) rather than because they spent a certain amount of time exploring the state space. The reward is likely to be non-Markovian as it depends of the trajectory a user follows and not just the last state he visits. Indeed, for a user trying to learn the content of a MOOC, a chapter is more rewarding when visited for the first time. 
An interesting direction for future work would be to tackle such challenging problem.

%
%
%
%

\end{document}